\documentclass[AMA,Times1COL]{WileyNJDv5}

\articletype{Research Article}

\usepackage{textcomp}
\usepackage{graphicx}
\usepackage{amsmath,amssymb,amsthm}
\usepackage{bm}
\usepackage{mathtools}
\usepackage{longtable,tabularx}
\usepackage{enumitem}
\usepackage{booktabs}
\usepackage{multirow}
\usepackage{makecell}
\usepackage{rotating}
\usepackage{float}
\usepackage{subcaption}
\usepackage{comment}
\usepackage{setspace}
\makeatletter
\let\njd@normalsize\normalsize
\def\normalsize{\@setfontsize\normalsize{10pt}{13pt}}
\expandafter\let\csname algorithm\endcsname\relax\expandafter\let\csname endalgorithm\endcsname\relax
\expandafter\let\csname algorithm*\endcsname\relax\expandafter\let\csname endalgorithm*\endcsname\relax
\expandafter\let\csname procedure\endcsname\relax\expandafter\let\csname endprocedure\endcsname\relax
\expandafter\let\csname procedure*\endcsname\relax\expandafter\let\csname endprocedure*\endcsname\relax
\expandafter\let\csname function\endcsname\relax\expandafter\let\csname endfunction\endcsname\relax
\expandafter\let\csname function*\endcsname\relax\expandafter\let\csname endfunction*\endcsname\relax

\usepackage[ruled,vlined,linesnumbered]{algorithm2e}
\let\normalsize\njd@normalsize
\makeatother
\SetKwInput{KwIn}{Input}
\SetKwInput{KwOut}{Output}
\SetKw{KwReturn}{return}

\SetAlgoCaptionLayout{algcapraggedright}


\usepackage[section]{placeins}

\usepackage{url}

\usepackage{etoolbox}
\makeatletter
\renewcommand{\operator@font}{\mathgroup\symmtoperatorfont}
\AtBeginEnvironment{algorithm}{\def\small{\fontsize{9}{11}\selectfont}}
\renewcommand\section{\@startsection{section}{1}{\z@}{-27pt \@plus -2pt \@minus -2pt}{12\p@}%
  {\fontsize{12}{14}\selectfont\bfseries\raggedright\hyphenpenalty\@M}}
\renewcommand\subsection{\@startsection{subsection}{2}{\z@}{-23pt \@plus -2pt \@minus -2pt}{12\p@}%
  {\fontsize{12}{14}\selectfont\bfseries\raggedright\hyphenpenalty\@M}}
\renewcommand\subsubsection{\@startsection{subsubsection}{3}{\z@}{-20pt \@plus -2pt \@minus -2pt}{12\p@}%
  {\fontsize{12}{14}\selectfont\bfseries\raggedright\hyphenpenalty\@M}}
\renewcommand{\bibfont}{\reset@font\fontfamily{\rmdefault}\fontsize{8.5}{11}\selectfont
  \baselineskip=10\p@\raggedright}
\makeatother
\historydates{}
\doiheadtext{}
\startpage{1}
\makeatletter
\def\oddfoot@titlepage@info{\vbox{\hsize\textwidth{\noindent\rule{\textwidth}{.5pt}\vskip-40pt
  {\fontsize{7}{8}\selectfont\hfill{\pagenumfont\@FirstPg}}\vskip10pt}}}
\makeatother
\hypersetup{hidelinks}

\begin{document}

\title{Constrained Extreme Gradient Boosting for Parametric Proper Orthogonal Decomposition Subspace Prediction}

\author[1]{Melika Baghi}
\author[1]{Xiao Liu}
\author[1]{Kamran Paynabar}

\authormark{BAGHI \textsc{et al.}}
\titlemark{CONSTRAINED EXTREME GRADIENT BOOSTING FOR PARAMETRIC PROPER ORTHOGONAL DECOMPOSITION SUBSPACE PREDICTION}

\address[1]{\orgdiv{H.~Milton Stewart School of Industrial and Systems Engineering},
\orgname{Georgia Institute of Technology}, \orgaddress{\state{Atlanta, GA 30332}, \country{USA}}}

\corres{Xiao Liu. \email{xliu976@gatech.edu}}

\abstract[Abstract]{High-fidelity simulations such as computational fluid dynamics and finite
element analysis are essential for modelling complex engineering systems but
remain prohibitively expensive for tasks including parametric studies,
optimisation and real-time control. Projection-based reduced-order models
(ROMs) alleviate this cost by projecting full-order dynamics onto
low-dimensional subspaces. On the other hand, projection-based ROMs can become
less robust under parameter variation, and one approach is to adapt the
projection basis (that is, the subspace) as the parameters change. We propose
constrained extreme gradient boosting (cXGBoost) for predicting the proper
orthogonal decomposition (POD) basis at given parameter settings. In
particular, given the observed parameter-subspace pairs, we map the subspaces
from the Grassmann manifold to a Euclidean space. Then, cXGBoost is
trained in this Euclidean space with a constraint that the norm of the predicted
vectors is bounded to ensure the injectivity of the mapping between the
Euclidean space and the Grassmann manifold. On four parametric problems, a
cylinder wake, Kolmogorov flow, the Kuramoto--Sivashinsky equation and a
forced beam, cXGBoost has lower mean reconstruction errors than existing
approaches, namely the projected Gaussian process and Grassmann
interpolation.}


\maketitle

\section{Introduction}

High-fidelity numerical solvers, such as Computational Fluid Dynamics (CFD) and Finite Element Analysis (FEA), are standard tools for the prediction, design, and control of complex physical systems. These simulations support applications such as aerodynamic shape optimisation \cite{amsallem2011} and structural health monitoring, resolving multiscale transport and structural phenomena with high spatial degrees of freedom.
The computational cost remains prohibitive for many-query tasks such as parametric design exploration, uncertainty quantification, model-predictive control, and digital twin operations, where dozens or hundreds of simulations may be required in rapid succession. For example, in aerodynamic design, evaluating a single geometry at multiple flow conditions consumes hours of compute time per configuration. In shock-dominated flow analysis, even one high-resolution run can take days on a cluster \cite{halder2022non}.

Reduced-Order Models (ROMs) have been widely adopted to mitigate this computational bottleneck. Among these, projection-based model reduction remains a popular solution \cite{Sun2024}. By projecting high-dimensional dynamics onto a low-dimensional subspace spanned by a small number of modes (or, bases), ROMs deliver orders-of-magnitude speedups. 
Examples include near real-time aerodynamic load prediction \cite{amsallem2011}, fast parametric studies of wake flows \cite{Sun2024}, efficient evaluation of thermal loads in reentry trajectories, hypersonic aeroheating analysis \cite{Ching2024}, data-driven aeroelastic prediction \cite{Shu2023}, partitioned structural dynamics simulations \cite{KimPO2024}, rotating-detonation-engine simulations \cite{Farcas2023RDE}, and aeroservoelastic--gust response analysis \cite{KimASEG2025}.

For parametric systems, the spanning basis needs to be sufficiently representative so that data generated from various parameter settings are well represented. 
For example, in the classical cylinder wake problems, the Proper Orthogonal Decomposition (POD)–Galerkin ROMs that perform well at moderate Reynolds numbers break down as the flow becomes more nonlinear and advective, requiring many more modes to capture essential dynamics and risking instability from truncated structures \cite{lee2020model}. In shock-dominated flows, linear subspaces struggle to capture discontinuities without additional stabilisation, often leading to spurious oscillations \cite{halder2022non}. 
To address the parameter dependence of POD bases, one approach, among others, is to perform basis interpolation \cite{amsallem2008interpolation,amsallem2011} with successful demonstrations in aeroelastic flutter prediction, aerodynamic shape optimisation, and structural dynamics. Refinements have addressed issues such as mode crossing and robustness to irregular parameter sampling \cite{Friderikos2022}. It is known that the space spanned by a POD basis is a point on the Grassmann manifold that consists of $r$-dimensional subspaces in $\mathbb{R}^n$ with $n$ and $r$ respectively being the original and reduced-order dimensions. Hence, a tangent space can be attached to a given base point on the Grassmann manifold, and the subspaces  can be projected from the Grassmann manifold to the tangent space through a logarithmic map. Interpolation can then be performed efficiently on the tangent space, and the interpolated vector on the tangent space is mapped back to the Grassmann manifold through an exponential map. 

Following this idea, \cite{liu2023tree} proposed the  regression tree on Grassmann manifold. A tree
is grown by splitting the tree node (i.e., a binary partition of the parameter space) to
maximise the Riemannian distance between two subspaces from the left and right daughter nodes. This approach divides the parameter space into sub-regions, and the POD basis for a sub-region is obtained using only the data generated from that sub-region. The method successfully demonstrated the use of supervised learning for adapting POD basis. Because the predictions generated by simple regression trees are non-smooth, a recent work proposed the projected Gaussian process (pGP) for adapting POD basis \cite{liu2025pGP}. In this method, a mapping is established between the Grassmann manifold and a Euclidean space where a Gaussian process (GP) can be constructed. Then, the constructed GP is wrapped onto the Grassmann manifold that enables the prediction of POD basis for given parameters. Other methods for adapting POD basis include the naive GP approach \cite{giovanis2020data}, and subspace angle interpolation \cite{ye2016schubert}.  We refer the readers to \cite{liu2025pGP,benner2015survey} for comprehensive surveys. 

In this paper, we propose a new approach, known as the constrained Extreme Gradient Boosting (cXGBoost), for predicting POD basis as parameters change. Here, the problem of adapting the POD basis is formulated as a supervised statistical learning problem, for which the goal is to learn a mapping from the parameter space to the Grassmann manifold that contains the subspaces. In particular, using the mapping between a (constrained) Euclidean space and the Grassmann manifold established in a recent work of \cite{liu2025pGP}, each subspace in the training dataset is mapped to a vector in a Euclidean space. This involves firstly mapping the subspace to a tangent space attached to a reference point of the Grassmann manifold, and then mapping the horizontal lift of the tangent vector to the target Euclidean space. Hence, it becomes possible to adapt the basic framework of XGBoost described in \cite{Chen2016} to learn the relationship between parameters and the vectors in the Euclidean space. 

As an ensemble-tree-based method, XGBoost is a widely used and efficient machine learning algorithm, which makes it worth investigating for the problem of adapting POD basis. Unlike kernel-based methods (e.g., Gaussian Processes) which typically enforce smoothness assumptions, decision trees naturally model non-smooth transitions and discontinuities. This characteristic makes the tree-based boosting framework particularly well-suited for problems where the optimal subspace may change abruptly across the parameter space. Gradient boosting also reduces the bias of a single decision tree while keeping its flexibility to capture non-linear dependencies.
For any new parameters, the predicted vectors are mapped back to the Grassmann manifold to obtain the predicted optimal subspaces. To ensure the injectivity of the mapping between the Euclidean space and the Grassmann manifold, a constraint must be imposed such that the norm of the predicted vectors stays below $\pi/2$ according to the latest result of \cite{liu2025pGP}. 
By integrating the manifold injectivity constraint directly into the leaf updates of the trees, where each leaf value solves a quadratically constrained quadratic program (QCQP) that reduces to a Euclidean projection, cXGBoost is proposed and described in detail in this paper. cXGBoost is compared with two existing methods on four parametric problems. On all four, cXGBoost has a lower mean reconstruction error than both baselines, although its margin over the pGP on Kolmogorov flow is small. A controlled ablation compares constrained and unconstrained boosting on the predicted coordinates, to isolate the effect and the limits of the training-time constraint.

The rest of the paper is organised as follows. Section~\ref{sec:method}
reviews POD and the Grassmann manifold, describes the Euclidean coordinate
and its norm bound, and presents cXGBoost. Section~\ref{sec:examples}
evaluates it on four problems, namely the flow around a cylinder, Kolmogorov flow,
the Kuramoto--Sivashinsky equation and a forced beam with two parameters. Section~\ref{sec:ablation} examines how
the constraint changes the boosting fit and how the predictions behave at
unseen parameters. Section~\ref{sec:conclusions} concludes the paper and
points to future work.
\section{Constrained Extreme Gradient Boosting for POD Subspace Prediction}
\label{sec:method}

\subsection{Proper Orthogonal Decomposition for Projection-Based Model Reduction}

We first present a quick review of POD and introduce some basic notation.  Consider a  system parameterised by a set of parameters $\bm{\theta} = (\theta_1, \theta_2, \ldots, \theta_d) \in \mathcal{P} \subset \mathbb{R}^d$. 
For each parameter $\bm{\theta}$, the solution of the system is a space-time field, $(t, s) \in [0, T] \times \mathcal{S} \;\mapsto\; x(t, s; \bm{\theta})$, 
where $\mathcal{S}$ is the spatial domain and $T>0$ is the time horizon.  
At discrete spatial locations $s_1,\dots,s_n$ and any time instance $t$, the solution (i.e., a snapshot at time $t$) can be collected in a vector $\bm{x}(t; \bm{\theta})
=(x(t,s_1;\bm{\theta}),\,x(t,s_2;\bm{\theta}),\,\ldots,\,x(t,s_n;\bm{\theta}))^{\!T}$. 
Combining all snapshots from times $t_1,t_2,\cdots,t_{n_T}$ yields the snapshot data matrix
\begin{equation}
\bm{D}(\bm{\theta})
=\big[\bm{x}(t_1;\bm{\theta}),\,\bm{x}(t_2;\bm{\theta}),\,\ldots,\,\bm{x}(t_{n_T};\bm{\theta})\big]
\in \mathbb{R}^{n\times n_T}.
\end{equation}

Projection-based model reduction seeks a low-dimensional subspace $V_r$ that minimises the error
\begin{equation}
\label{eq:J}
\mathcal{J}(V_r)
=\sum_{i=1}^{n_T}\!\big\|
\bm{x}(t_i;\bm{\theta})
-\pi_{V_r}\big(\bm{x}(t_i;\bm{\theta})\big)
\big\|^2
\end{equation}
where $\pi_{V_r}$ denotes the orthogonal projection onto $V_r$ and $||\cdot||=\sqrt{\left \langle \cdot,\cdot \right \rangle}$ and $\left \langle \cdot,\cdot \right \rangle$ is the inner product of the Hilbert space. If the subspace $V_r$ is spanned by a matrix basis $\bm{\Phi}_{\bm{\theta}}=[\bm{\phi}_1,\ldots,\bm{\phi}_r]\in\mathbb{R}^{n\times r}$ with orthonormal columns, the matrix basis is found by $\min_{\bm{\Phi}_{\bm{\theta}}}\;
\|
\bm{D}(\bm{\theta})
-\bm{\Phi}_{\bm{\theta}} \bm{\Phi}_{\bm{\theta}}^{\!T}\bm{D}(\bm{\theta})
\|_F^2$
with $\|\cdot\|_F$ being the Frobenius norm.  
By the Eckart–Young theorem, the optimal basis is formed by the first $r$ left singular vectors obtained through the thin Singular Value Decomposition (SVD) of the snapshot data matrix, $\bm{D}(\bm{\theta})
=\bm{U}\bm{\Sigma}\bm{V}^{\!T}$, i.e., $\bm{\Phi}_{\bm{\theta}}
=(\bm{u}_1,\,\bm{u}_2,\,\ldots,\,\bm{u}_r)$ which is known as the POD basis. 

It is seen that the optimal POD basis, obtained from the snapshot data generated at the parameter setting $\bm{\theta}$, is only optimal to represent the data generated from that parameter setting $\bm{\theta}$ by minimising (\ref{eq:J}). 
Hence, for parametric dynamical systems, 
it is critical for the POD modes to be sufficiently representative so that data generated from various parameter settings are well represented. One possible approach is to adapt the POD basis $\bm{\Phi}_{\bm{\theta}}$ as the parameter changes. 

\subsection{The Proposed cXGBoost} \label{sec:mapping}
Suppose that the snapshot data matrices are generated at a set of parameter settings, $\bm{\Lambda}=\{\bm{\theta}_1, \bm{\theta}_2, \cdots, \bm{\theta}_k\}$, and the optimal POD bases $\bm{\Phi}_1, \bm{\Phi}_2, \cdots, \bm{\Phi}_k \in \mathbb{R}^{n\times r} $ are respectively obtained for each parameter in $\bm{\Lambda}$. 
Then, the POD basis prediction problem involves finding the POD basis $\bm{\Phi}^*$ at a new parameter setting $\bm{\theta}^*\notin  \bm{\Lambda}$. In this subsection, we show how cXGBoost can be used to learn the relationship between parameters and POD bases.  

A POD basis $\bm{\Phi}$ is an $n\times r$ matrix with orthonormal columns, and the collection of $n\times r$ matrices with orthonormal columns forms a compact Stiefel manifold denoted by $\mathcal{ST}(r,n)$. Each matrix basis $\bm{\Phi}$ spans an $r$-dimensional linear subspace in $\mathbb{R}^n$, and the collection of such subspaces forms a differentiable manifold known as the Grassmann manifold, $\mathcal{G}(r,n):=\{\mathbb{W}\subset \mathbb{R}^n, \mathrm{dim}(\mathbb{W})=r\}$. Note that, if $\bm{\Phi}$ spans a subspace $p \in \mathcal{G}(r,n)$, so does $\bm{\Phi}\bm{Q}$ for any $r\times r$ matrix $\bm{Q}$ such that $\bm{Q}^T\bm{Q}=\bm{I}_r$. Therefore, there exists a fiber bundle $\pi: \bm{\Phi} \in \mathcal{ST}(r,n) \mapsto \pi(\bm{\Phi}) = p \in \mathcal{G}(r,n)$, 
which indicates that any point $p \in \mathcal{G}(r,n)$ can be represented by a point of the fiber $\pi^{-1}(p)$ \cite{Friderikos2022}. 

To establish the Riemannian metric on $\mathcal{G}(r,n)$, a unique tangent space $\mathcal{T}_p\mathcal{G}(r,n)$ is attached to a point $p \in \mathcal{G}(r,n)$. The tangent space $\mathcal{T}_p\mathcal{G}(r,n)$ has the same dimension of $\mathcal{G}(r,n)$ (but is isomorphic to $\mathbb{R}^{r\times (n-r)}$, i.e., $\mathcal{T}_p\mathcal{G}(r,n) \simeq \mathbb{R}^{r\times (n-r)}$) and is equipped with a scalar product. For any point $p \in \mathcal{G}(r,n)$ and the attached tangent space $\mathcal{T}_p\mathcal{G}(r,n)$, the mapping between $p$ and $v \in \mathcal{T}_p\mathcal{G}(r,n)$ is established through the exponential and logarithm maps. In particular, the exponential map is defined as
\begin{equation} \label{eq:ExpMap}
\text{Exp}_p:  v \in \mathcal{T}_p\mathcal{G}(r,n) \mapsto \pi(\bm{\Phi} \bm{V}_{\bm{Z}}\cos(\bm{\Sigma}_{\bm{Z}}) + \bm{U}_{\bm{Z}}\sin(\bm{\Sigma}_{\bm{Z}}))\in\mathcal{G}(r,n)
\end{equation}
where $\bm{\Phi} \in \pi^{-1}(p)$ and $\bm{Z}=\bm{U}_{\bm{Z}}\bm{\Sigma}_{\bm{Z}}\bm{V}_{\bm{Z}}^T$ is the SVD of $\bm{Z} \in \mathbb{H}_{\bm{\Phi}}$. Note that, for a given $\Phi$ such that $\pi(\Phi)=p$, the tangent space $\mathcal{T}_p\mathcal{G}(r,n)$ is isomorphic to the horizontal space $\mathbb{H}_{\bm{\Phi}}$. For any $v \in \mathcal{T}_p\mathcal{G}(r,n)$, there is a unique $\bm{Z} \in \mathbb{H}_{\bm{\Phi}}$ known as the horizontal lift of $v$ \cite{Friderikos2022}. 

For any point $p \in \mathcal{G}(r,n)$ and an open set $U_{\bm{p}}:=\{p' \in \mathcal{G}(r,n)\}$, the logarithm map is defined as
\begin{equation} \label{eq:LogMap}
    \text{Log}_p: p' \in U_{\bm{p}} \mapsto \text{Log}_p(p') \in \mathcal{T}_p\mathcal{G}(r,n). 
\end{equation}

The tangent space $\mathcal{T}_p\mathcal{G}(r,n)$ is equipped with a scalar product.
For any two velocity vectors, $v_1, v_2 \in \mathcal{T}_p\mathcal{G}(r,n)$, we have 
\begin{equation}
    \left \langle v_1,v_2 \right \rangle := \left \langle \bm{Z}_1,\bm{Z}_2 \right \rangle = \mathrm{trace}(\bm{Z}_1^T \bm{Z}_2)
\end{equation}
where $\bm{Z}_1$ and $\bm{Z}_2$ are $n\times r$ matrices from the horizontal space of $\bm{\Phi} \in \pi^{-1}(p)$ denoted by $\mathbb{H}_{\bm{\Phi}}:=\{\bm{Z}\in \mathbb{R}^{n\times r}; \bm{Z}^T\bm{\Phi}=0\}$. 

There is an isomorphism between the horizontal space $\mathbb{H}_{\bm{\Phi}}$ and the tangent space $\mathcal{T}_p\mathcal{G}(r,n)$, i.e., $\pi_{\mathbb{H}}: \mathbb{H}_{\bm{\Phi}} \rightarrow  \mathcal{T}_p\mathcal{G}(r,n)$. Hence, 
it is possible to find a unique $\bm{Z} \in \mathbb{H}_{\bm{\Phi}}$ such that $\pi_{\mathbb{H}}(\bm{Z}) = v \in  \mathcal{T}_p\mathcal{G}(r,n)$, 
and the matrix $\bm{Z}$ is called the \textit{horizontal lift} of the vector $v \in  \mathcal{T}_p\mathcal{G}(r,n)$. 
The existence of such an isomorphism may suggest that a matrix-variate regression model can be employed to model the relationship between $\bm{\theta}$ and the matrix $\bm{Z}$. 

However, as pointed out by \cite{liu2025pGP}, this seemingly natural approach needs to be carefully carried out because the matrix $\bm{Z}$ should be defined on the horizontal space $\mathbb{H}_{\bm{\Phi}}$ given a base point $p$, rather than the Euclidean space $\mathbb{R}^{n\times r}$. To ensure that the predicted $\bm{Z}$ by the regression model belongs to the horizontal space  $\mathbb{H}_{\bm{\Phi}}$, \cite{liu2025pGP} introduced an injective mapping between the horizontal space  $\mathbb{H}_{\bm{\Phi}}$ and a Euclidean space $\mathbb{R}^{nr-r}$, and a regression model can be used to establish the mapping from the parameter space to the $\mathbb{R}^{nr-r}$. 

To elaborate, for a given basepoint $p \in \mathcal{G}(r,n)$ and its corresponding basis matrix $\bm{\Phi}=[\bm{\phi}_{1},\cdots,\bm{\phi}_{r}]$, i.e., $\pi(\bm{\Phi})=p$, it is possible to obtain an $n\times r$ matrix $\bm{Z}$ from a $(nr-r)\times nr$ matrix $\bm{F}$  and a $(nr-r)\times 1$ vector $\bm{y}$ as follows
\begin{equation} \label{eq:y_to_Z}
\bm{Z} = \mathrm{Mat}_{n, r}(\bm{F}^{T}\bm{y})
\end{equation} 
such that $\bm{Z} $ belongs to the horizontal space $\mathbb{H}_{\bm{\Phi}}$, i.e.,  $\bm{Z}^T \bm{\Phi} = \bm{0}$. Here, the matrix $\bm{F}$ has orthonormal rows and satisfies the conditions  $\tilde{\bm{\Phi}}\bm{F}^T=\bm{0}$ and $\bm{F}\bm{F}^T=\bm{I}$, where $\tilde{\bm{\Phi}}=\mathrm{diag}\{ \bm{\phi}_{1}^T,\bm{\phi}_{2}^T,\cdots,\bm{\phi}_{r}^T\}$ is an $r\times nr$ matrix with orthonormal rows. 

Hence, once a regression model (i.e., cXGBoost in this paper) has been constructed, one could predict a vector $\bm{y} \in \mathbb{R}^{nr-r}$, and the predicted vector can be mapped back to the Grassmann manifold $\mathcal{G}(r,n)$ through 
\begin{equation}
\mathcal{P}_p:  \bm{y}\in\mathbb{R}^{nr-r} \mapsto \pi(\bm{\Phi} \bm{V}_{\bm{Z}}\cos(\bm{\Sigma}_{\bm{Z}}) + \bm{U}_{\bm{Z}}\sin(\bm{\Sigma}_{\bm{Z}}))\in\mathcal{G}(r,n). 
\label{eq:mapping}
\end{equation}
Here, $\mathcal{P}_p := \text{Exp}_p \circ \tilde{\mathcal{P}}_p$ where $\tilde{\mathcal{P}}_p$ maps an $(nr-r)$-dimensional vector to the tangent space $\mathcal{T}_p\mathcal{G}(r,n)$, 
\begin{equation}
\tilde{\mathcal{P}}_p:  \bm{y}\in\mathbb{R}^{nr-r} \mapsto v \in \mathcal{T}_p\mathcal{G}(r,n).
\end{equation}
As shown in \cite{liu2025pGP}, this mapping is injective on the open set $\Omega_p:=\left\{\bm{y} \in  \mathbb{R}^{nr-r}; ||\bm{y}||_2 < \frac{\pi}{2} \right\}$.

\vspace{6pt}
Hence, the proposed cXGBoost learns the relationship between parameters and the vector $\bm{y} \in \mathbb{R}^{nr-r}$, given a chosen reference point $p\in\mathcal{G}(r,n)$. Because $||\bm{y}||_2$ needs to be smaller than $\frac{\pi}{2}$, a constraint needs to be incorporated into the algorithm of cXGBoost. 
In particular, given a training dataset $\mathcal{D} = \{(\bm{\theta}_i, \bm{y}_i)\}_{i=1}^N$, where $\bm{\theta}_i \in \mathbb{R}^d$ and $\bm{y}_i \in \mathbb{R}^{nr-r}$, 
an ensemble of $K$ additive regression trees is given by
\begin{equation}
\hat{\bm{y}}_i^{(K)} = \sum_{k=1}^K f_k(\bm{\theta}_i),
\qquad f_k \in \mathcal{F},
\end{equation}
where $f(\bm{\theta})$ is a binary tree with $J$ leaves (terminal nodes), 
$\mathcal{F}
=\{ f(\bm{\theta}) = \bm{w}_{q(\bm{\theta})} |  
q:\mathbb{R}^{d}\!\to\!\{1,\dots,J\},
\bm{w}\in\mathbb{R}^{J\times(nr-r)} \}$
denotes the space of regression trees, and $q$ represents a mapping from the input 
$\bm{\theta}$ to its corresponding leaf index given the tree structure. 
Suppose that a number of $k-1$ trees have been grown, the $k$-th tree is added to minimise the following objective function
\begin{equation}
\begin{split}
\mathcal{L}^{(k)} = & 
\sum_{i=1}^{N} \ell (\bm{y}_i,\, \hat{\bm{y}}_i^{(k-1)} + f_k(\bm{\theta}_i))
+ \Psi(f_k) \\    
& s.t. \quad \|\hat{\bm{y}}_i^{(k-1)} + f_k(\bm{\theta}_i)\|_2 < \frac{\pi}{2} 
\quad \text{for }i=1,2,\cdots,N.
\end{split}
\end{equation}

Here, a quadratic loss function $\ell (\cdot,\cdot)$ is adopted, the penalty term $\Psi(f_k)=\sum_{j=1}^{J}\tfrac{1}{2}\lambda\|\bm{w}_j\|_2^2 + \gamma J$ regularises the model complexity, and the constraint is needed to ensure the injectivity of the mapping $\mathcal{P}_p$ defined in (\ref{eq:mapping}).  
The Taylor expansion of the loss function around $\hat{\bm{y}}_i^{(k-1)}$ yields
\begin{equation}
\ell(\bm{y}_i, \hat{\bm{y}}_i^{(k-1)} + f_k(\bm{\theta}_i)) \approx \ell(\bm{y}_i, \hat{\bm{y}}_i^{(k-1)}) + \bm{g}_i^\top f_k(\bm{\theta}_i) + \frac{1}{2} f_k(\bm{\theta}_i)^\top \bm{h}_i f_k(\bm{\theta}_i)
\end{equation}
where
 $\bm{g}_i = \nabla_{\hat{\bm{y}}_i^{(k-1)}} \ell(\bm{y}_i, \hat{\bm{y}}_i^{(k-1)})$ and $\bm{h}_i = \nabla^2_{\hat{\bm{y}}_i^{(k-1)}} \ell(\bm{y}_i, \hat{\bm{y}}_i^{(k-1)})$ are respectively the gradient vector and
 Hessian matrix. Removing the constant term $\ell(\bm{y}_i, \hat{\bm{y}}_i^{(k-1)})$, we obtain a simpler form of the approximated objective function for growing the $k$-th tree: 
 \begin{equation}
\tilde{\mathcal{L}}^{(k)} =  
\sum_{i=1}^{N} \bm{g}_i^\top f_k(\bm{\theta}_i) + \frac{1}{2} f_k(\bm{\theta}_i)^\top \bm{h}_i f_k(\bm{\theta}_i)
+ \Psi(f_k).
\label{eq:objective_tree}
\end{equation}

Given a specific tree structure with $J$ leaves, let $I_j = \{i | q(\bm{\theta}_i) = j \}$ be the instance set of leaf $j=1,2,\cdots,J$, one could re-write the objective function (\ref{eq:objective_tree}) as
\begin{equation}
\tilde{\mathcal{L}}^{(k)} =  \sum_{j=1}^{J} \left\{\bm{G}_j^\top \bm{w}_j + \frac{1}{2} \bm{w}_j^\top (\bm{H}_j+\lambda \bm{I}) \bm{w}_j\right\} + \gamma J
\end{equation}
where
$\bm{G}_j = \sum_{i \in I_j} \bm{g}_i$ and 
$\bm{H}_j = \sum_{i \in I_j} \bm{h}_i$. 
Hence, for any candidate tree structure for the $k$-th tree, finding the optimal weights $\bm{w}_j$ on leaf $j=1,2,\cdots,J$ involves solving a quadratically constrained quadratic program (QCQP), which Section~\ref{sec:cxgboost} states in \eqref{eq:leafqcqp} and solves.

\subsection{Geometric Norm Constraint and Coordinate Reduction}
\label{sec:whyconstrained}
\label{sec:reduction}

The logarithm map \eqref{eq:LogMap} is computed by the algorithm of Bendokat et al.~\cite{bendokat2024handbook}, which needs no matrix inverse and remains defined for principal angles up to $\pi/2$. Write $\mathbb{S}$ for the set of coordinates $\bm{y}=\bm{F}\,\mathrm{vec}(\bm{Z})$ of horizontal matrices $\bm{Z}\in\mathbb{H}_{\bm{\Phi}}$, a linear subspace of $\mathbb{R}^{nr-r}$ of dimension $r(n-r)$ that contains every training target. For a coordinate $\bm{y}\in\mathbb{S}$, let
$\bm{Z} = \operatorname{Mat}_{n,r}(\bm{F}^{\top}\bm{y})$ be its horizontal
tangent matrix. As long as the largest singular value of $\bm{Z}$, its spectral
norm $\|\bm{Z}\|_2$, stays below $\pi/2$, the singular values of $\bm{Z}$ are
the principal angles between the subspace returned by \eqref{eq:mapping} and
the reference, and the Grassmann exponential map is one-to-one
\cite{bendokat2024handbook}. This region is larger than the ball $\|\bm{y}\|_2<\pi/2$ of Section~\ref{sec:mapping}. From here on $\Omega_p$ denotes this larger region,
\begin{equation}
\Omega_p = \bigl\{\, \bm{y} \in \mathbb{S} \;:\; \|\bm{Z}\|_2 < \tfrac{\pi}{2} \,\bigr\}.
\label{eq:injdomain}
\end{equation}
The coordinate construction of Section~\ref{sec:mapping} preserves the norm,
$\|\bm{y}\|_2 = \|\bm{Z}\|_F$, because $\bm{F}$ has orthonormal rows and $\mathrm{vec}(\bm{Z})$ lies in its row space for a horizontal $\bm{Z}$. Since $\|\bm{Z}\|_2 \le \|\bm{Z}\|_F$ for every
matrix,
\begin{equation}
\|\bm{y}\|_2 \le R < \frac{\pi}{2}
\;\;\Longrightarrow\;\;
\|\bm{Z}\|_2 \le \|\bm{y}\|_2 < \frac{\pi}{2}.
\label{eq:frobimpliesspec}
\end{equation}
Lying in the Euclidean ball $\|\bm{y}\|_2 \le R$ is therefore a sufficient, although
conservative, condition for membership of $\Omega_p$. This ball is used
because the projection onto it is radial and has a closed form, and because it remains a ball after the coordinate reduction described next. It is closed, with $R=\pi/2-10^{-9}$, so the projection onto it lies inside
$\Omega_p$.

A last preliminary concerns the size of the coordinate. For large problems the coordinate $\bm{y}$, with $nr-r$ entries, is
high-dimensional. Principal component analysis (PCA) therefore reduces the training
coordinates to at most $c=20$ components, computed on the training data alone,
before the regression model is fitted. The cap of twenty was fixed in advance
and not tuned. Write $\bar{\bm{y}}$ for the mean of the training
coordinates and $\bm{V}$ for the retained components, so that a reduced
coordinate $\tilde{\bm{y}}$ stands for $\bm{y}=\bar{\bm{y}}+\bm{V}\tilde{\bm{y}}$.
Since $\bm{V}$ has orthonormal columns, the bound $\|\bm{y}\|_2\le R$ becomes
\begin{equation}
\bigl\|\tilde{\bm{y}} + \bm{V}^{\top}\bar{\bm{y}}\bigr\|_2^{2}
\;\le\; R^{2}
- \bigl\|\bar{\bm{y}} - \bm{V}\bm{V}^{\top}\bar{\bm{y}}\bigr\|_2^{2},
\label{eq:pcaball}
\end{equation}
so the feasible set in the reduced coordinate is again a Euclidean ball,
centred at $-\bm{V}^{\top}\bar{\bm{y}}$. The ball is
nonempty whenever the part of the training mean outside the retained
components has norm at most $R$. The implementation checks this condition
before fitting. When the reduction is used, the leaf optimisation and boosting below work on $\tilde{\bm{y}}$ and the ball \eqref{eq:pcaball}, called $\mathcal{C}$ below, while the prediction-time safeguard still acts on the full coordinate $\bm{y}$. Without the reduction, $\mathcal{C}$ is the ball of radius $R$ about the origin.

\subsection{Implementation of cXGBoost}
\label{sec:cxgboost}

cXGBoost regresses the Euclidean coordinate of Section~\ref{sec:mapping} and requires every leaf value of every tree to keep the accumulated training predictions inside the ball of Section~\ref{sec:whyconstrained}. If the ordinary XGBoost leaf value already does so, it is kept. Otherwise it is replaced by a feasible constrained value, as described below. Boosting with vector-valued leaves has been studied before \cite{zhang2021gbdtmo,cortes2026vectorleaf}, but without a constraint that ties the leaf value to the accumulated prediction.

Let $\mathcal{D} = \{(\bm{\theta}_i, \bm{y}_i)\}_{i=1}^N$ be the training set, with
$\bm{\theta}_i \in \mathbb{R}^d$ and $\bm{y}_i \in \mathbb{S}\subset\mathbb{R}^{nr-r}$. An ensemble of $K$
additive regression trees predicts
\begin{equation}
\hat{\bm{y}}_i^{(K)} = \bm{y}^{(0)} + \sum_{k=1}^K \eta\, f_k(\bm{\theta}_i),
\qquad f_k \in \mathcal{F},
\end{equation}
where each $f_k$ is a binary tree with $J$ leaves. The space of such trees is $\mathcal{F} =\{ f(\bm{\theta}) = \bm{w}_{q(\bm{\theta})} \mid q:\mathbb{R}^{d}\!\to\!\{1,\dots,J\}, \bm{w}\in\mathbb{R}^{J\times(nr-r)} \}$, where $q$ maps the input $\bm{\theta}$ to its leaf index and $\bm{w}$ holds one value per leaf. The
learning rate $\eta \in (0,1]$ shrinks the contribution of each tree, so the
accumulated prediction advances as
$\hat{\bm{y}}_i^{(k)} = \hat{\bm{y}}_i^{(k-1)} + \eta\, f_k(\bm{\theta}_i)$
from a base score $\hat{\bm{y}}_i^{(0)} = \bm{y}^{(0)}$, the mean of the training
targets, which is projected radially onto the ball of radius $R$ if it lies
outside. As in ordinary shrinkage boosting, the $k$-th tree is fitted as a full
update and the learning rate is applied afterwards. Suppose that $k-1$ trees have been grown.
The $k$-th tree is added to minimise the objective
\begin{equation}
\begin{split}
\mathcal{L}^{(k)} = & 
\sum_{i=1}^{N} \ell (\bm{y}_i,\, \hat{\bm{y}}_i^{(k-1)} + f_k(\bm{\theta}_i))
+ \Psi(f_k) \\    
& \text{s.t.} \quad \|\hat{\bm{y}}_i^{(k-1)} + f_k(\bm{\theta}_i)\|_2 \le R 
\quad \text{for }i=1,2,\ldots,N.
\end{split}
\label{eq:constrained_objective}
\end{equation}

Here $\ell(\bm{y}_i,\hat{\bm{y}}_i)=\tfrac{1}{2}\|\bm{y}_i-\hat{\bm{y}}_i\|_2^2$ is the squared Euclidean loss, the penalty $\Psi(f_k)=\sum_{j=1}^{J}\tfrac{1}{2}\lambda\|\bm{w}_j\|_2^2 + \gamma J$ regularises the model complexity, and the constraint keeps the accumulated training predictions in the region where the map $\mathcal{P}_p$ of \eqref{eq:mapping} is one-to-one.
Since $\ell$ is quadratic, its second-order Taylor expansion around $\hat{\bm{y}}_i^{(k-1)}$ has no remainder,
\begin{equation}
\ell(\bm{y}_i, \hat{\bm{y}}_i^{(k-1)} + f_k(\bm{\theta}_i)) = \ell(\bm{y}_i, \hat{\bm{y}}_i^{(k-1)}) + \bm{g}_i^\top f_k(\bm{\theta}_i) + \frac{1}{2} f_k(\bm{\theta}_i)^\top \bm{h}_i f_k(\bm{\theta}_i),
\end{equation}
where
 $\bm{g}_i = \nabla_{\hat{\bm{y}}_i^{(k-1)}} \ell(\bm{y}_i, \hat{\bm{y}}_i^{(k-1)})$ and $\bm{h}_i = \nabla^2_{\hat{\bm{y}}_i^{(k-1)}} \ell(\bm{y}_i, \hat{\bm{y}}_i^{(k-1)})$ are the gradient vector and the
 Hessian matrix. Removing the constant term $\ell(\bm{y}_i, \hat{\bm{y}}_i^{(k-1)})$, we obtain the objective for growing the $k$-th tree,
 \begin{equation}
\tilde{\mathcal{L}}^{(k)} =  
\sum_{i=1}^{N} \bm{g}_i^\top f_k(\bm{\theta}_i) + \frac{1}{2} f_k(\bm{\theta}_i)^\top \bm{h}_i f_k(\bm{\theta}_i)
+ \Psi(f_k).
\label{eq:objective}
\end{equation}

Fix a tree structure with $J$ leaves and let $I_j = \{i : q(\bm{\theta}_i) = j \}$ be
the set of training parameters that reach leaf $j$. The objective
(\ref{eq:objective}) then separates over leaves,
\begin{equation}
\tilde{\mathcal{L}}^{(k)} =  \sum_{j=1}^{J} \left\{\bm{G}_j^\top \bm{w}_j + \frac{1}{2} \bm{w}_j^\top (\bm{H}_j+\lambda \bm{I}) \bm{w}_j\right\} + \gamma J,
\label{eq:leafobjective}
\end{equation}
where
$\bm{G}_j = \sum_{i \in I_j} \bm{g}_i$ and 
$\bm{H}_j = \sum_{i \in I_j} \bm{h}_i$. 
Hence, for any candidate structure of the $k$-th tree, the optimal value $\bm{w}_j$ of leaf $j=1,2,\ldots,J$ solves the quadratically constrained quadratic program
\begin{equation}
\begin{split}
\min_{\bm{w}_j} \quad &
\bm{G}_j^\top \bm{w}_j + 
\frac{1}{2}\,\bm{w}_j^\top(\bm{H}_j + \lambda \bm{I})\bm{w}_j \\ 
& \text{s.t.} \quad
\bm{w}_j^\top\bm{w}_j + 2\bm{w}_j^\top \hat{\bm{y}}_i^{(k-1)}
+ \|\hat{\bm{y}}_i^{(k-1)}\|_2^2
- R^{2} \leq 0, \\
& \qquad\qquad\quad \text{for all } i \in I_j.
\end{split}
\label{eq:leafqcqp}
\end{equation}

Equations \eqref{eq:constrained_objective} and \eqref{eq:leafqcqp} are written for the ball of radius $R$ about the origin. With the reduction of Section~\ref{sec:reduction}, $R$ and the origin are replaced by the radius and the centre of $\mathcal{C}$ in \eqref{eq:pcaball}, and nothing else below changes, because $\mathcal{C}$ is also a Euclidean ball.
The objective is strictly convex and the feasible set is convex, so the
solution is unique. Under this loss every Hessian $\bm{h}_i$ is the
identity, so $\bm{H}_j+\lambda \bm{I}$ is a multiple of the identity. The solution is then the Euclidean projection of the ordinary XGBoost leaf value
$-(\bm{H}_j+\lambda \bm{I})^{-1}\bm{G}_j$ onto the intersection of the balls $\mathcal{B}_i=\{\bm{w}:\hat{\bm{y}}_i^{(k-1)}+\bm{w}\in\mathcal{C}\}$, one for each training parameter $i$ in the leaf. If a single ball determines the projection, it has the closed form $\Pi_i(\bm{w})=\Pi_{\mathcal{C}}(\hat{\bm{y}}_i^{(k-1)}+\bm{w})-\hat{\bm{y}}_i^{(k-1)}$, where $\Pi_{\mathcal{C}}$ is the radial projection onto $\mathcal{C}$. Otherwise the projection is computed by Dykstra's algorithm \cite{dykstra1983algorithm,boyle1986dykstra}, and the iterate is checked against every ball of the leaf, to a tolerance that the margin of $R$ below $\pi/2$ absorbs. If the iterate still fails the check after all violated balls have been included, it is scaled towards zero by the largest factor in $[0,1]$ that satisfies every ball. Zero is always feasible, because every accumulated prediction satisfies the bound, so this scaled iterate is feasible but in general not the optimal leaf value. The check, the radial projection and each Dykstra sweep take time linear in the number of balls times the coordinate dimension, so no general-purpose optimiser is needed. Algorithm~\ref{alg:leaf} states the leaf update step by step.

The bound in \eqref{eq:leafqcqp} applies to the fully updated prediction $\hat{\bm{y}}_i^{(k-1)}+\bm{w}_j$, while boosting moves the prediction only to $\hat{\bm{y}}_i^{(k-1)}+\eta\bm{w}_j$. The base score is feasible by construction, and $\hat{\bm{y}}_i^{(k-1)}$ is feasible by induction, so the new prediction
\begin{equation}
\hat{\bm{y}}_i^{(k-1)} + \eta\, \bm{w}_j
= (1-\eta)\,\hat{\bm{y}}_i^{(k-1)}
+ \eta \left( \hat{\bm{y}}_i^{(k-1)} + \bm{w}_j \right)
\label{eq:shrinkfeas}
\end{equation}
is a convex combination of two points inside the ball and lies inside it for
any $\eta \in (0,1]$. The accumulated prediction therefore satisfies the bound
at every training parameter and at every $k$. 

Every prediction also stays in the subspace $\mathbb{S}$ of Section~\ref{sec:mapping}, so it maps back to a horizontal tangent matrix. The base score and every leaf value are linear combinations of the training targets and the accumulated predictions, which all lie in $\mathbb{S}$. With the reduction of Section~\ref{sec:reduction} the same holds, because $\bar{\bm{y}}$ and the columns of $\bm{V}$ lie in $\mathbb{S}$.

The $k$-th tree is grown by the usual greedy search, with the gain of a candidate split taken as the decrease in the leaf terms of \eqref{eq:leafobjective} under these leaf values, and a split kept when its gain exceeds $\gamma$.

The guarantee covers the training parameters, and for a scalar parameter it extends to every parameter. Every split point lies midway between two consecutive training values, so the split points of all the trees cut the parameter axis into intervals that each contain a training parameter. A new parameter therefore takes the same leaf in every tree as a training parameter in its interval and receives that parameter's prediction, which lies inside the ball. With two or more parameters, split points on different parameters can enclose a region with no training parameter, and a new parameter there can select a combination of leaves across trees that no training parameter selects. A predicted coordinate whose norm exceeds $R$ is therefore projected radially onto the ball before it is mapped back to the
Grassmann manifold, the same prediction-time safeguard as the pGP uses \cite{liu2025pGP}.

\begin{algorithm}[!t]
\small
\caption{Constrained Extreme Gradient Boosting (cXGBoost)}
\label{alg:cXGBoost}
\SetAlgoLined
\DontPrintSemicolon
\KwIn{$\{(\bm{\theta}_i,\bm{\Phi}_i)\}_{i=1}^{N}$, a reference subspace $p$ and the settings $c$, $R$, $K$, $\eta$, $D$, $\lambda$, $\gamma$, $n_{\min}$}
\KwOut{the predicted subspace at a new parameter $\bm{\theta}$}
Compute the target coordinates $\bm{y}_i$ of the training bases (Section~\ref{sec:mapping})\;
Reduce them by PCA to $\tilde{\bm{y}}_i$, with mean $\bar{\bm{y}}$ and at most $c$ components $\bm{V}$, and form the ball $\mathcal{C}$ of \eqref{eq:pcaball}\;
Set the base score $\bm{y}^{(0)}$ to the mean of the $\tilde{\bm{y}}_i$, projected radially onto $\mathcal{C}$, and $\hat{\bm{y}}_i\leftarrow\bm{y}^{(0)}$ for all $i$\;
\For{$k=1,\dots,K$}{
  $\bm{g}_i\leftarrow\hat{\bm{y}}_i-\tilde{\bm{y}}_i$ for all $i$\;
  Grow the tree $f_k$ greedily to depth at most $D$, keeping a split when its gain exceeds $\gamma$ and each child holds at least $n_{\min}$ training parameters, with every leaf value from Algorithm~\ref{alg:leaf}\;
  $\hat{\bm{y}}_i\leftarrow\hat{\bm{y}}_i+\eta\,f_k(\bm{\theta}_i)$ for all $i$\;
}
At a new $\bm{\theta}$, set $\hat{\bm{y}}(\bm{\theta})=\bar{\bm{y}}+\bm{V}\bigl(\bm{y}^{(0)}+\sum_{k=1}^{K}\eta\,f_k(\bm{\theta})\bigr)$, project it radially onto the ball of radius $R$ if it lies outside, and return $\mathcal{P}_p\bigl(\hat{\bm{y}}(\bm{\theta})\bigr)$\;
\end{algorithm}

\begin{algorithm}[!t]
\small
\caption{Leaf value under the norm constraint, \texttt{LeafValue}$(I)$}
\label{alg:leaf}
\SetAlgoLined
\DontPrintSemicolon
\SetKwInput{KwNote}{Notation}
\KwIn{the training parameters $I$ of the leaf, $\{\bm{g}_i,\hat{\bm{y}}_i\}_{i\in I}$, the ball $\mathcal{C}$ of centre $\bm{a}$ and radius $\rho$, and $\lambda$}
\KwOut{a leaf value $\bm{w}\in\bigcap_{i\in I}\mathcal{B}_i$, to the tolerance of the feasibility check}
\KwNote{$\mathcal{B}_i=\{\bm{w}:\hat{\bm{y}}_i+\bm{w}\in\mathcal{C}\}$ and $\Pi_i(\bm{w})=\Pi_{\mathcal{C}}(\hat{\bm{y}}_i+\bm{w})-\hat{\bm{y}}_i$}
$\bm{w}\leftarrow-\sum_{i\in I}\bm{g}_i\,/\,(|I|+\lambda)$\;
$A\leftarrow\{i\in I:\bm{w}\notin\mathcal{B}_i\}$\;
\lIf{$A=\emptyset$}{\Return $\bm{w}$}
\ForEach{$i\in A$}{
  $\bm{w}'\leftarrow\Pi_i(\bm{w})$\;
  \lIf{$\bm{w}'\in\mathcal{B}_j$ for all $j\in I$}{\Return $\bm{w}'$}
}
\Repeat{$A$ stops growing}{
  $\bm{w}'\leftarrow$ Dykstra's projection of $\bm{w}$ onto $\bigcap_{i\in A}\mathcal{B}_i$, stopped when the relative change of the iterate falls below $10^{-10}$ or after $200$ sweeps\;
  $A'\leftarrow\{i\in I:\|\hat{\bm{y}}_i+\bm{w}'-\bm{a}\|_2^2>\rho^2+10^{-9}\}$\;
  \lIf{$A'=\emptyset$}{\Return $\bm{w}'$}
  $A\leftarrow A\cup A'$\;
}
$s\leftarrow\max\{s\in[0,1]:s\bm{w}'\in\mathcal{B}_i\text{ for all }i\in I\}$\;
\Return $s\,\bm{w}'$\;
\end{algorithm}
\section{Numerical Examples}
\label{sec:examples}

This section investigates the performance of cXGBoost and compares it with two existing methods on four examples, the flow around a cylinder, Kolmogorov flow, the Kuramoto--Sivashinsky equation and parametric beam dynamics.

\subsection{Settings of cXGBoost}
\label{sec:settings}

The settings that vary by example, the number of boosting rounds $K$, the
learning rate $\eta$ and the maximum depth $D$ of each tree, are given with
each example. Where an example has a fixed training split, they were chosen by
leave-one-out cross-validation over the training parameters of that split, on a grid of depths $2$, $3$
and $4$, learning rates $0.1$, $0.3$ and $0.5$, and $50$, $150$ and $300$
boosting rounds. The other two examples were tuned on the same grid by four-fold cross-validation within each training fold of the five-fold evaluation, and each then uses, in all five folds, the setting that most folds selected. Because every example uses one setting in all five folds, the error estimates may be optimistic. All analyses of this section, including the fixed-size study at the end of it, use these settings.
The remaining settings are shared. A leaf holds at least $n_{\min} = 2$ training parameters, and the penalty weights are $\gamma = 10^{-3}$ and $\lambda = 10^{-2}$. For each component of the parameter, up to twenty split points are tried, spread evenly over the midpoints between consecutive training values.
No row or feature subsampling is used, so the fit is deterministic for a given training set and the only randomness in the cross-validation is the fold partition, drawn once with the fixed seed $42$.

\subsection{Baselines}
\label{sec:baselines}

cXGBoost is compared with Grassmann interpolation
\cite{amsallem2008interpolation} and the projected Gaussian process (pGP)
\cite{liu2025pGP}. Both work in the same tangent space of the Grassmann manifold as cXGBoost. The pGP introduced the coordinate of Section~\ref{sec:mapping}, and its treatment of
the bound is what cXGBoost changes, while interpolation is the established
tangent-space reference. Predictors with another representation of the
subspace, such as Gaussian process subspace prediction \cite{zhang2022gps},
would confound the regression
mechanism with the representation, so the comparison is restricted to these
two.

Grassmann interpolation \cite{amsallem2008interpolation} lifts the training
bases to the tangent space, interpolates the tangent matrices and maps the result back. The norm of the interpolated tangent matrix is not
bounded by $\pi/2$.

The projected Gaussian process \cite{liu2025pGP} regresses the Euclidean
coordinate of Section~\ref{sec:mapping}, the same coordinate used by cXGBoost,
with one Gaussian process per output. All outputs share one squared-exponential kernel, with an amplitude and one length scale per parameter component. Its hyperparameters are fitted by maximising the summed log marginal likelihood from three starting points. The parameters are standardised, and $10^{-8}$ is added to the kernel diagonal. Its
posterior mean is unconstrained, and a prediction whose coordinate lies
outside the Euclidean ball is projected radially onto the ball before it is
mapped back to the Grassmann manifold. Every error reported for the pGP is the
error of that delivered prediction.

\FloatBarrier

\label{sec:protocol}

Reconstruction accuracy is measured by the relative error of a predicted basis
$\hat{\bm{\Phi}}$ on the snapshots at the corresponding parameter,
\begin{equation}
e(\bm{\theta}) = \frac{\|\bm{D}(\bm{\theta})
 - \hat{\bm{\Phi}}\hat{\bm{\Phi}}^{\!\top}\bm{D}(\bm{\theta})\|_F}
{\|\bm{D}(\bm{\theta})\|_F}.
\label{eq:relerr}
\end{equation}
This measures how well the predicted subspace represents the held-out snapshots, not the accuracy of a reduced-order model integrated in time. Unless stated otherwise,
reconstruction errors are obtained by five-fold cross-validation over the full
parameter sweep, with the parameters assigned to the five folds at random,
each held-out parameter scored once, and the distribution over held-out parameters reported. The box plots that
show these distributions mark the quartiles, with whiskers at $1.5$ times the
interquartile range. Lower error is better throughout. Because every predictor is scored at
the same held-out parameters, the comparison is also made parameter by parameter, counting the
parameters at which cXGBoost has the lower error. The cylinder and the beam also have a fixed training split, stated with each example, which is used to select the settings of cXGBoost and to fit any single model that is shown. For
each fit the reference subspace of Section~\ref{sec:mapping} is selected from
the training set alone, by a rule that is not tuned, and is shared by all
methods compared within that fit. Each scalar sweep is ordered by its
parameter and the reference is the subspace of the training parameter at the
median position. The coordinate of a POD basis relative to that reference is
its target coordinate. Table~\ref{tab:examples} gives the parameter range,
the data and the split of each example.

\begin{table}[htbp]
\centering
\caption{The four examples and the size of each cross-validation fit.}
\label{tab:examples}
\footnotesize
\setlength{\tabcolsep}{4pt}
\begin{tabular}{lcccc}
\toprule
 & Cylinder & Kolmogorov & \makecell{Kuramoto--\\Sivashinsky} & Beam \\
\midrule
Parameter & $\mathrm{Re}$ & $\mathrm{Re}$ & $L$ & $(\mu_1, \mu_2)$ \\
Range & $[10, 1200]$ & $[8, 45]$ & $[12, 55]$ & $[3, 9]^2$ \\
Parameters in the sweep & $46$ & $34$ & $44$ & $25$ \\
Snapshots per parameter & $2{,}000$ & $200$ & about $500$ & $100$ \\
Spatial points, $n$ & $93{,}600$ & $1{,}024$ & $64$ & $200$ \\
POD rank, $r$ & $5$ & $6$ & $8$ & $10$ \\
Training parameters per fit & $36$ or $37$ & $27$ or $28$ & $35$ or $36$ & $20$ \\
Test parameters per fit & $10$ or $9$ & $7$ or $6$ & $9$ or $8$ & $5$ \\
Fixed split, training and test & $23$ and $23$ & none & none & $16$ and $9$ \\
\bottomrule
\end{tabular}
\end{table}

The POD rank is held fixed within each sweep and across all compared methods,
because $\mathcal{G}(r,n)$ contains subspaces of fixed dimension $r$ and a
parameter-dependent rank selected by an energy threshold would place the bases
on different Grassmann manifolds. A fixed rank does not retain the same share
of snapshot energy at every parameter, and truncation to rank $r$ introduces
an irreducible reconstruction component shared by all three predictors. cXGBoost and the pGP use the PCA reduction of Section~\ref{sec:reduction}, refitted within each training fold, whereas interpolation works on the tangent matrices directly. The reduced ball of \eqref{eq:pcaball} is nonempty in
every training fold.

\subsection{Example I: Flow Around a Cylinder}
\label{sec:cylinder}

The cylinder example is a broad Reynolds number sweep, from a steady wake to vortex shedding. It is the two-dimensional
incompressible flow past a circular cylinder in a channel of $520 \times 180$ lattice units, with a cylinder of
radius $20$ centred at $(130, 90)$, so the blockage ratio is the same at every Reynolds number. A uniform inflow of $0.04$ lattice units per time step enters at the left, carrying a sinusoidal perturbation of relative size $10^{-4}$ that starts the vortex shedding. A Zou--He velocity condition \cite{zou1997construction} holds at the inlet, the outlet is a zero-gradient outflow, the top and bottom are periodic, and bounce-back on the cylinder enforces no slip. The flow is solved by the lattice Boltzmann method on the D2Q9 lattice
with the Bhatnagar--Gross--Krook collision operator
\cite{bhatnagar1954model,qian1992lattice}. The Reynolds number is formed on the
cylinder radius, $\mathrm{Re} = U a/\nu$ with $a = 20$ and $U$ the inlet velocity, and is set through the viscosity, the only quantity that changes across the sweep. Above $\mathrm{Re} = 160$ a Prandtl mixing-length correction \cite{prandtl1925turbulenz} stands in for the unresolved dissipation. At every time step it lowers the relaxation frequency by an amount computed from the field-wide magnitude of the non-equilibrium stress, with mixing length $0.22\,D_h$, where $D_h = 2a = 40$ is the cylinder diameter, so the correction is uniform in space and varies in time.
The simulation is two-dimensional over the whole sweep, so at the higher Reynolds numbers it is a parametric test problem for subspace prediction rather than a model of the physical wake.
Each simulation runs for $20{,}000$ time steps from a uniform initial state.

\begin{figure}[!htb]
    \centering
    \includegraphics[width=\linewidth]{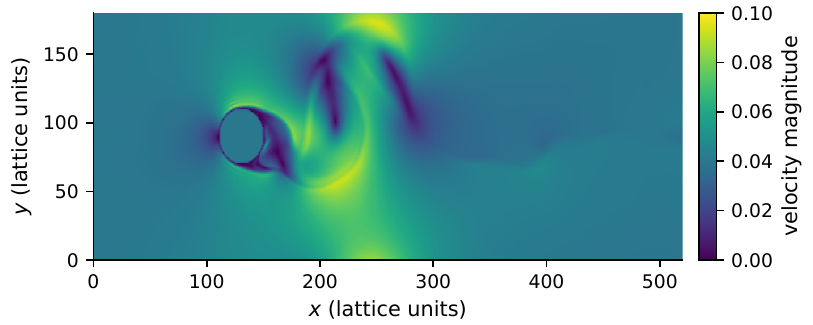}
    \caption{Velocity magnitude of the cylinder wake at $\mathrm{Re}=1000$ after $16{,}000$ time steps, one snapshot of the record.}
    \label{fig:setup_cyl_field}
\end{figure}

At each Reynolds number the velocity magnitude on the whole $520 \times 180$ lattice, $n = 93{,}600$ points, is recorded every ten time steps, and the resulting $2{,}000$ columns form the snapshot matrix. The record used for the snapshots
is the whole run, including the start-up transient. A rank-$5$ POD basis is
computed per parameter, so each Reynolds number contributes one point of
$\mathcal{G}(5, 93{,}600)$. The basis keeps at least $98.7\%$ of the snapshot energy
at every Reynolds number. The full coordinate has $nr - r = 467{,}995$
entries, and the reduction to twenty components keeps between $98.7\%$ and
$99.1\%$ of its energy in each training fold. Figure~\ref{fig:setup_cyl_field} shows one of the
snapshots.

\FloatBarrier

The sweep has $46$ Reynolds numbers in $[10, 1200]$. At the lowest of them
the wake is steady rather than shedding, and the low end of the sweep is far
from the reference subspace. At $\mathrm{Re} = 1233$, $1267$, $1333$ and
$1500$ the solver did not remain finite over the run, so the sweep ends at
$1200$. The fixed training
split takes alternate Reynolds numbers, $23$ for training and $23$ for
testing, and is used to select the settings of cXGBoost and to produce the field reconstructions below. Leave-one-out cross-validation over its $23$ training Reynolds numbers selects $K = 300$
boosting rounds, a learning rate $\eta = 0.1$ and a maximum depth $D = 4$.

The mean relative errors are $0.085$ for cXGBoost, $0.096$ for the pGP and
$0.383$ for interpolation, with medians of $0.076$, $0.091$ and $0.197$.
cXGBoost has a lower error than interpolation at $37$ of the $46$ Reynolds
numbers and than the pGP at $35$ of the $46$. 

\begin{figure}[!htb]
    \centering
    \includegraphics[width=0.9\linewidth]{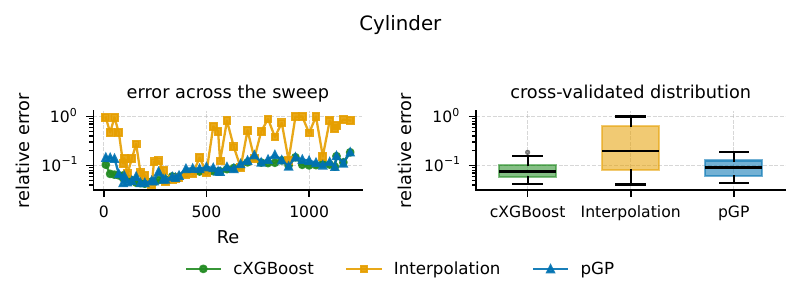}
    \caption{Cylinder. The relative error \eqref{eq:relerr} at each Reynolds number and its cross-validated distribution.}
    \label{fig:cylinder_cmp}
\end{figure}

Interpolation is closest to the other two over $\mathrm{Re} \in [300, 500)$,
with a mean error of $0.069$, and degrades on both sides of that range, to
$0.27$ below $\mathrm{Re}=300$, $0.49$ over $\mathrm{Re} \in [500, 1100)$ and
$0.74$ for $\mathrm{Re} \ge 1100$. The errors of cXGBoost and the pGP vary much less across the sweep (Figure~\ref{fig:cylinder_cmp}).

\begin{figure}[!htb]
    \centering
    \includegraphics[width=0.8\linewidth]{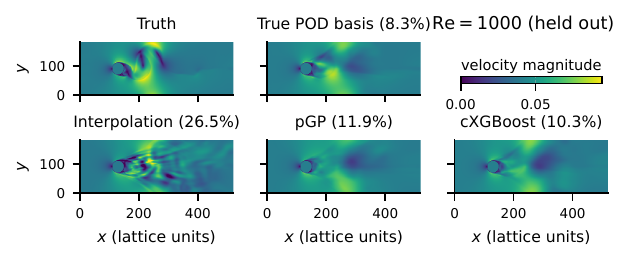}\\[2pt]
    \includegraphics[width=0.8\linewidth]{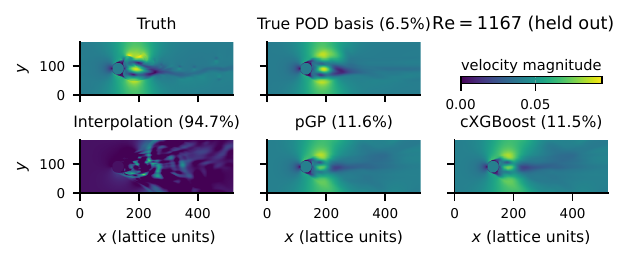}
    \caption{Velocity magnitude at two held-out Reynolds numbers after $16{,}000$ time steps, true and projected onto the true POD basis and onto each predicted subspace, with the relative error of each over the whole record.}
    \label{fig:field_mod}
\end{figure}

Figure~\ref{fig:field_mod} compares reconstructed velocity fields at two
held-out Reynolds numbers. At $\mathrm{Re}=1000$, interpolation has a
reconstruction error of $26.5\%$, compared with $11.9\%$ for the pGP and
$10.3\%$ for cXGBoost. At $\mathrm{Re}=1167$ the corresponding errors are
$94.7\%$, $11.6\%$ and $11.5\%$. The POD basis of each held-out record,
which no five-dimensional subspace can improve on, has errors of $8.3\%$ and
$6.5\%$, so a large part of the error of cXGBoost and the pGP is the truncation to
rank $5$ that every predictor shares. The larger interpolation error is
visible as distortion of the late-time vortex structure. 

\FloatBarrier
\subsection{Example II: Kolmogorov Flow}
\label{sec:kolmogorov}

As the Reynolds number increases, the laminar shear solution loses stability and the flow goes through a sequence of bifurcations into sustained vortex dynamics \cite{platt1991investigation}, so its dominant coherent
structures reorganise over the sweep. The flow is driven by a monochromatic body force and
governed by the two-dimensional incompressible Navier--Stokes equations on a
doubly periodic square domain. In vorticity form,
\begin{equation}
\frac{\partial \omega}{\partial t} + J(\psi, \omega) = \nu \nabla^2 \omega + f,
\qquad \nabla^2 \psi = -\omega,
\qquad f = -n_f\cos(n_f\, y),
\end{equation}
where $\omega$ is the vorticity, $\psi$ the streamfunction, $J(\psi,\omega)=\psi_x\omega_y-\psi_y\omega_x$ the Jacobian carrying the nonlinear advection and $n_f = 4$ the forcing wavenumber. The Reynolds number $\mathrm{Re} = 1/\nu$ indexes the sweep.

The solver is pseudospectral in space on a $64 \times 64$ grid with two-thirds
dealiasing, and time stepping uses a Lawson integrating-factor Runge--Kutta scheme of
order four \cite{lawson1967generalized} at $\Delta t = 2.5\times 10^{-3}$, so the stiff viscous term is
integrated analytically, as in the exponential integrators of Cox and Matthews~\cite{cox2002exponential}.

\begin{figure}[!htb]
    \centering
    \includegraphics[width=0.86\linewidth]{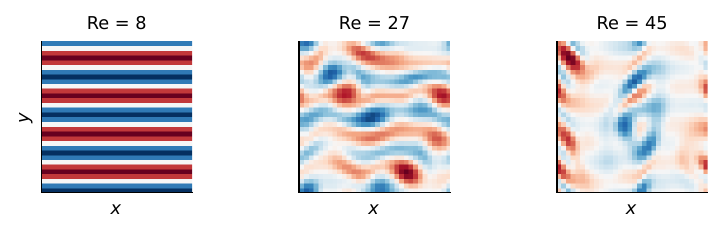}
    \caption{Kolmogorov vorticity at three Reynolds numbers of the sweep.}
    \label{fig:setup_kol}
\end{figure}

Forty time units are discarded as transient and the next twenty are sampled
as $200$ snapshots of the vorticity field, which are stored on every second
grid point, giving $n = 1{,}024$. A rank-$6$ POD basis is computed at each
Reynolds number, giving points of $\mathcal{G}(6, 1{,}024)$. It retains less of
the snapshot energy where the solution is most complex, with a median of
$96.3\%$ falling to $83.5\%$ at the worst Reynolds number, and the reduction to
twenty components keeps between $84.0\%$ and $87.3\%$ of the coordinate energy
in each training fold.
At low Reynolds number the $n_f = 4$ forcing gives eight laminar bands, and Figure~\ref{fig:setup_kol} shows them breaking up along the sweep.

The sweep has $34$ evenly spaced Reynolds numbers in $[8, 45]$. There is no fixed training split. Four-fold cross-validation within each training fold selects $K = 50$ boosting rounds, a learning rate $\eta = 0.1$ and a maximum depth $D = 2$ in all five folds.

\begin{figure}[!htb]
    \centering
    \includegraphics[width=0.9\linewidth]{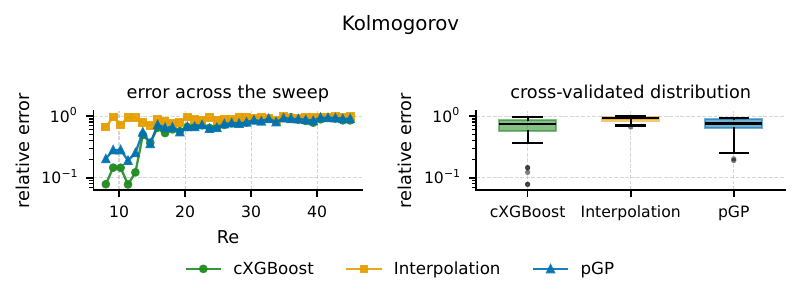}
    \caption{Kolmogorov. Error at each Reynolds number and its cross-validated distribution.}
    \label{fig:kolmogorov_cmp}
\end{figure}

The mean relative errors are $0.663$ for cXGBoost, $0.695$ for the pGP and $0.894$ for interpolation. cXGBoost has a lower error than interpolation at $33$ of the $34$ Reynolds numbers and than the pGP at $21$ of the $34$. Figure~\ref{fig:kolmogorov_cmp} shows that cXGBoost and the pGP
perform best at the laminar end of the sweep, with mean errors of $0.21$ and $0.30$ below
$\mathrm{Re}=15$. Their errors approach the interpolation error as the flow
becomes turbulent, and none of the three reconstructs the flow accurately once sustained vortex dynamics dominate. Refitting at $r = 12$
more than halves the irreducible reconstruction component and lowers the mean error of every
predictor by between $1.4\%$ and $3.0\%$ of its value, so a higher rank alone does not fix the reconstruction.

\begin{figure}[!htb]
    \centering
    \includegraphics[width=0.95\linewidth]{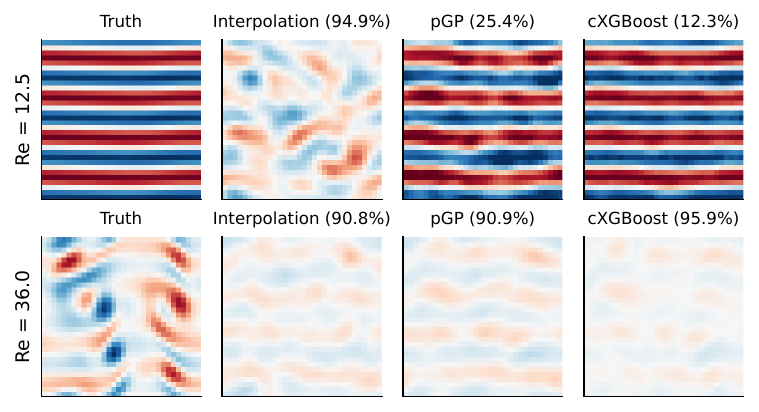}
    \caption{Kolmogorov vorticity at a laminar and a turbulent held-out Reynolds number, true and projected onto each predicted subspace, with the relative error of each.}
    \label{fig:kolmogorov_field}
\end{figure}

Figure~\ref{fig:kolmogorov_field} compares the reconstructed vorticity at two
held-out Reynolds numbers, each with the models of the fold that scored it. At
$\mathrm{Re} = 12.5$ cXGBoost and the pGP recover the laminar bands, with errors of $12.3\%$ and $25.4\%$, whereas interpolation does not, at $94.9\%$.
At $\mathrm{Re} = 36.0$ none of the three recovers the vortices, with errors of
$90.8\%$ for interpolation, $90.9\%$ for the pGP and $95.9\%$ for cXGBoost.

\FloatBarrier
\subsection{Example III: Kuramoto--Sivashinsky Equation}
\label{sec:ks}

The Kuramoto--Sivashinsky equation is a one-dimensional test of
spatiotemporal chaos. On a periodic domain of length $L$ it reads
\begin{equation}
u_t = -u\,u_x - u_{xx} - u_{xxxx}, \qquad x \in [0, L).
\end{equation}
The domain length is the varying
parameter. Increasing $L$ admits progressively more unstable modes, and along the sweep the long-time solution switches repeatedly between steady cellular, travelling, time-periodic and chaotic states \cite{hyman1986kuramoto,kevrekidis1990back}. Each regime change alters which spatial
structures carry the energy, so the subspace often changes sharply from one domain length to the next.

The initial condition is $u(x,0)=\cos(2\pi x/L)\,\bigl(1+\sin(2\pi x/L)\bigr)$.
The equation is solved by a Fourier pseudospectral method with two-thirds dealiasing and advanced in time by a Lawson integrating-factor Runge--Kutta scheme of order four
\cite{lawson1967generalized}. The step is $0.5/(6\,k_{\max})$, the advective limit at an assumed amplitude of six, with $k_{\max}$ the
largest resolved wavenumber, which gives steps between $0.0047$ and $0.0091$ across the sweep. The
grid is chosen per run to hold the spacing at or below $0.35$, so the resolution
varies with $L$, from $64$ to $256$ points.

\label{sec:ks_data}

\begin{figure}[!htb]
    \centering
    \includegraphics[width=0.86\linewidth]{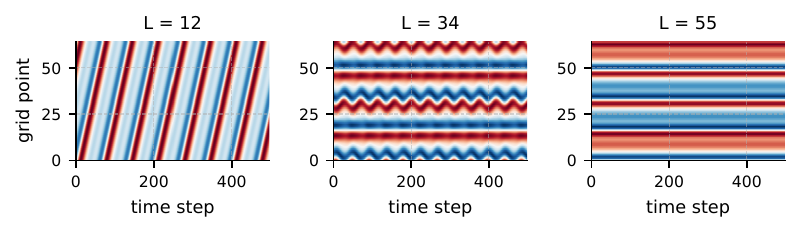}
    \caption{Kuramoto--Sivashinsky solution in space and time across the sweep.}
    \label{fig:setup_ks}
\end{figure}

The first $250$ time units are discarded as transient, and the next $250$ are sampled every half time unit,
giving about $500$ snapshots per run. Because the bases of a sweep must share the
number of spatial points $n$, every run is restricted
spectrally to the coarsest common grid, $n = 64$ equally spaced points in the normalised coordinate $x/L$, which keeps only the Fourier modes that every run resolves. A rank-$8$ POD basis is
computed at each domain length, giving points of $\mathcal{G}(8, 64)$. It
retains less of the snapshot energy where the solution is most complex, with a
median of $100\%$ falling to $81.0\%$ at the worst domain length, and the
reduction to twenty components keeps between $86.6\%$ and $91.5\%$ of the
coordinate energy in each training fold. At ten of the domain lengths with a steady solution, the snapshot matrix has numerical rank below eight, so the remaining POD modes are set by round-off.
Figure~\ref{fig:setup_ks} shows the solution at three domain lengths.

The sweep has $44$ domain lengths, the integers in $[12, 55]$. There is no fixed training split. Four-fold cross-validation within each training fold selects $K = 50$ boosting rounds, a learning rate $\eta = 0.1$ and a maximum depth $D = 3$ in four of the five folds, and this setting is used in all five.

\begin{figure}[!htb]
    \centering
    \includegraphics[width=0.9\linewidth]{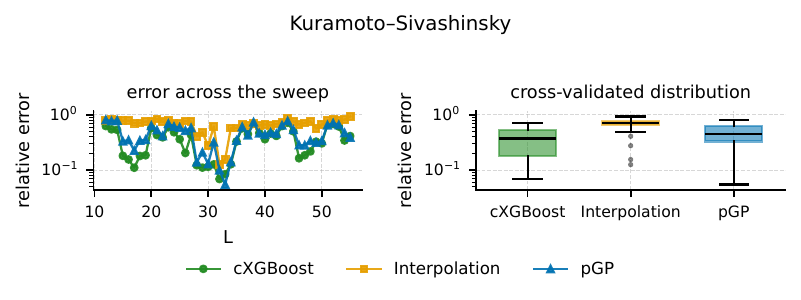}
    \caption{Kuramoto--Sivashinsky. Error at each domain length and its cross-validated distribution.}
    \label{fig:ks_cmp}
\end{figure}

The mean relative
errors are $0.371$ for cXGBoost, $0.455$ for the pGP and $0.684$ for interpolation. cXGBoost has a lower error than interpolation at all $44$ domain lengths and than the pGP at $38$ of the $44$, but its mean error is still large. The error of all three methods drops near $L = 33$, at the end of a range of steady solutions from $L = 26$ to $33$
(Figure~\ref{fig:ks_cmp}). Repeating the cross-validation without the ten domain lengths whose snapshot matrices have numerical rank below eight keeps the order of the three methods, with mean errors of $0.472$ for cXGBoost, $0.500$ for the pGP and $0.739$ for interpolation. cXGBoost is then below the pGP at $22$ of the $34$.

\begin{figure}[!htb]
    \centering
    \includegraphics[width=0.95\linewidth]{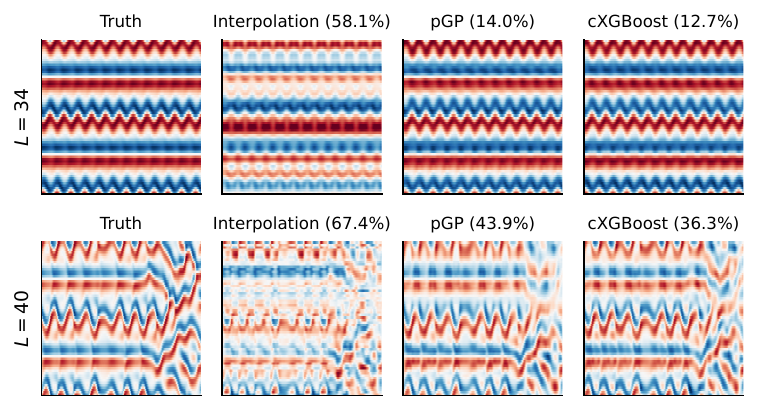}
    \caption{Kuramoto--Sivashinsky solution at a time-periodic and a chaotic held-out domain length, true and projected onto each predicted subspace, with the relative error of each.}
    \label{fig:ks_field}
\end{figure}

Figure~\ref{fig:ks_field} compares the reconstructed solution at two held-out
domain lengths, each with the models of the fold that scored it. At $L = 34$,
the time-periodic state of Figure~\ref{fig:setup_ks}, cXGBoost and the pGP
recover the oscillating cells, with errors of $12.7\%$ and $14.0\%$, whereas
interpolation does not, at $58.1\%$. At $L = 40$, in the chaotic range, the
errors are $36.3\%$ for cXGBoost, $43.9\%$ for the pGP and $67.4\%$ for
interpolation.

\FloatBarrier
\subsection{Example IV: Parametric Beam Dynamics}
\label{sec:beam}

The beam example is the transverse vibration of a damped Euler--Bernoulli beam on the unit interval $[0,1]$ with simply supported ends. The two forcing frequencies vary independently, so it tests cXGBoost with two parameters. The
displacement $u(x,t)$ satisfies
\begin{equation}
u_{tt} + 2\zeta u_t + \alpha^4 u_{xxxx}
= \sin(\mu_1 \pi t)\, g(x; 0.25) + \sin(\mu_2 \pi t)\, g(x; 0.75),
\end{equation}
with $\alpha = 0.25$ and $\zeta = 0.05$ held fixed. The forcing is a pair of
localised Gaussian loads $g(x;x_0)=\exp\{-\tfrac{1}{2}((x-x_0)/0.04)^2\}$ of
unit peak at $x_0 = 0.25$ and $x_0 = 0.75$, and the parameters
$\mu_1$ and $\mu_2$ are the angular frequencies of the two loads in units of
$\pi$, each taking values in $[3, 9]$.

The beam starts at rest. The equation is discretised on $200$ spatial points by a five-point finite difference for the fourth derivative, with the simply supported ends imposed through antisymmetric ghost points. It is advanced to $T = 8$ by an explicit central difference in time, with a step of $0.3\,\Delta x^{2}/\alpha^{2}$, about $1.2\times10^{-4}$, where $\Delta x$ is the grid spacing.

\begin{figure}[!htb]
    \centering
    \includegraphics[width=0.86\linewidth]{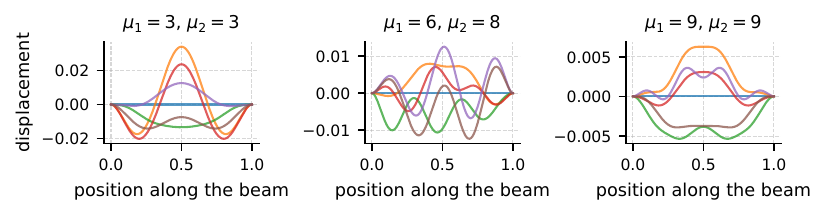}
    \caption{Beam displacement across the parameter grid, six time instants each.}
    \label{fig:setup_beam}
\end{figure}

At each parameter pair, displacement snapshots are recorded at $100$ equally
spaced times over the whole run. A rank-$10$ POD basis is
computed at each parameter pair, giving points of $\mathcal{G}(10, 200)$. It keeps more than $99.99\%$ of the snapshot energy at every parameter pair, and the reduction to twenty components is lossless.
Figure~\ref{fig:setup_beam} shows the response at three points of the grid.

The sweep has $25$ parameter pairs $(\mu_1, \mu_2)$ of integer frequencies
in $[3, 9]$, the $4 \times 4$ grid in which both are odd and the $3 \times 3$
grid in which both are even. The fixed training split is the $16$ odd-frequency pairs, with the $9$
even-frequency pairs held out, and is used to select the settings of
cXGBoost. Leave-one-out cross-validation over its $16$ training pairs selects $K = 50$ boosting
rounds, a learning rate $\eta = 0.1$ and a maximum depth $D = 4$.
The pairs are ordered with the $16$ odd-frequency pairs first and the $9$ even-frequency
pairs after them, each block ordered by $\mu_1$ and then $\mu_2$, and the
reference subspace of Section~\ref{sec:mapping} is the subspace of the
training pair at the median position of that ordering.

\begin{figure}[!htb]
    \centering
    \includegraphics[width=0.9\linewidth]{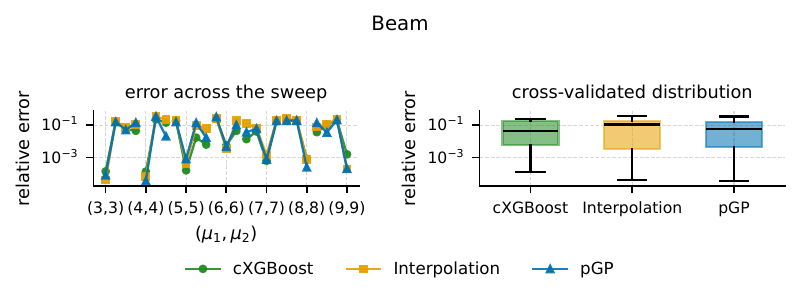}
    \caption{Beam. Error at each parameter pair, ordered by $\mu_1$ then $\mu_2$, and its cross-validated distribution.}
    \label{fig:beam_cmp}
\end{figure}

The mean relative errors are $0.084$ for cXGBoost, $0.100$ for the pGP and $0.119$ for interpolation, with medians of $0.044$, $0.059$ and $0.109$. cXGBoost has a lower error than interpolation at $20$ of the $25$ parameter pairs and than the pGP at $15$ of the $25$. The three error distributions are closer on the beam than on the other three examples (Figure~\ref{fig:beam_cmp}). Unlike the scalar sweeps, the beam also tests predictions at parameter pairs that select combinations of leaves no training pair selects, which the ablation study examines.

\FloatBarrier
\subsection{Comparison Across Examples}
\label{sec:examples_summary}

Table~\ref{tab:cv_error_stats} summarises the five-fold cross-validation results for the four examples. cXGBoost has the
lowest mean error on all four examples and a lower error than
interpolation at $134$ of the $149$ held-out parameters. Against the pGP it has the lower error at $35$ of the $46$ cylinder parameters, $21$ of the $34$ Kolmogorov parameters, $38$ of the $44$ Kuramoto--Sivashinsky parameters and $15$ of the $25$ beam parameter pairs. Within the five-fold partition, cXGBoost has the lowest mean in each of the five folds on the cylinder and Kuramoto--Sivashinsky and in four of the five on Kolmogorov and the beam.

\begin{table}[htbp]
\centering
\caption{Relative error under five-fold cross-validation over the full sweep.}
\label{tab:cv_error_stats}
\footnotesize
\setlength{\tabcolsep}{8pt}
\begin{tabular}{lccccc}
\toprule
Predictor & Parameters & Mean & Std & Median & Max \\
\midrule
\multicolumn{6}{l}{\emph{Cylinder}} \\
\quad cXGBoost & 46 & 0.0850 & 0.0329 & 0.0762 & 0.1854 \\
\quad Interpolation & 46 & 0.3833 & 0.3353 & 0.1970 & 0.9984 \\
\quad pGP & 46 & 0.0958 & 0.0384 & 0.0913 & 0.1874 \\
\quad \emph{cXGBoost lower at} & \multicolumn{5}{l}{\makebox[0pt][l]{$37$ of $46$ against interpolation, $35$ of $46$ against the pGP}} \\
\midrule
\multicolumn{6}{l}{\emph{Kolmogorov}} \\
\quad cXGBoost & 34 & 0.6626 & 0.2647 & 0.7445 & 0.9590 \\
\quad Interpolation & 34 & 0.8942 & 0.0859 & 0.9281 & 0.9872 \\
\quad pGP & 34 & 0.6951 & 0.2293 & 0.7541 & 0.9425 \\
\quad \emph{cXGBoost lower at} & \multicolumn{5}{l}{\makebox[0pt][l]{$33$ of $34$ against interpolation, $21$ of $34$ against the pGP}} \\
\midrule
\multicolumn{6}{l}{\emph{Kuramoto--Sivashinsky}} \\
\quad cXGBoost & 44 & 0.3706 & 0.1976 & 0.3775 & 0.7228 \\
\quad Interpolation & 44 & 0.6841 & 0.1680 & 0.7153 & 0.9375 \\
\quad pGP & 44 & 0.4550 & 0.2022 & 0.4464 & 0.8098 \\
\quad \emph{cXGBoost lower at} & \multicolumn{5}{l}{\makebox[0pt][l]{$44$ of $44$ against interpolation, $38$ of $44$ against the pGP}} \\
\midrule
\multicolumn{6}{l}{\emph{Beam}} \\
\quad cXGBoost & 25 & 0.0837 & 0.0854 & 0.0440 & 0.2471 \\
\quad Interpolation & 25 & 0.1195 & 0.1005 & 0.1090 & 0.3547 \\
\quad pGP & 25 & 0.1002 & 0.1000 & 0.0591 & 0.3417 \\
\quad \emph{cXGBoost lower at} & \multicolumn{5}{l}{\makebox[0pt][l]{$20$ of $25$ against interpolation, $15$ of $25$ against the pGP}} \\
\bottomrule
\end{tabular}
\end{table}

Raising the PCA cap of Section~\ref{sec:reduction} to $40$, which keeps every
component since no training fold has more than $37$ parameters, leaves the
cXGBoost mean at $0.085$ on the cylinder and moves it from $0.663$ to $0.671$ on Kolmogorov and from $0.371$ to $0.369$ on Kuramoto--Sivashinsky, so keeping every component gives no consistent gain.

To assess the predictors with fewer training simulations, each is
also trained on a fixed number of parameter settings and tested on the rest, $24$ Reynolds numbers and $22$ for the
cylinder, $24$ and $10$ for Kolmogorov, $6$ and $38$ for
Kuramoto--Sivashinsky, and $16$ parameter pairs and $9$ for the beam. Every reported value is the mean over eight independent draws of the split.

The three predictors keep the order of Table~\ref{tab:cv_error_stats}, with cXGBoost lowest on all four examples
(Table~\ref{tab:headline}). The margin is the reduction
$(e_{\mathrm{b}} - e_{\mathrm{c}})/e_{\mathrm{b}}$ in per cent, where
$e_{\mathrm{c}}$ is the mean error of cXGBoost and $e_{\mathrm{b}}$ that of
whichever published predictor has the lower error at that training size.

\begin{table}[htbp]
\centering
\caption{Mean relative error at a fixed number of training parameters, tested on the rest, with the standard deviation over the eight draws in parentheses.}
\label{tab:headline}
\footnotesize
\setlength{\tabcolsep}{4pt}
\begin{tabular}{lccccc}
\toprule
Example & Trained on & cXGBoost & Interpolation & pGP & Margin \\
\midrule
Cylinder & $24$ & 0.0890 & 0.2847 & 0.1003 & 11.3\% \\
 & & (0.0049) & (0.0666) & (0.0121) & \\[2pt]
Kolmogorov & $24$ & 0.6896 & 0.8994 & 0.7328 & 5.9\% \\
 & & (0.0356) & (0.0305) & (0.0470) & \\[2pt]
Kuramoto--Sivashinsky & $6$ & 0.4603 & 0.6284 & 0.5495 & 16.2\% \\
 & & (0.0718) & (0.0809) & (0.0700) & \\[2pt]
Beam & $16$ & 0.1082 & 0.1331 & 0.1239 & 12.7\% \\
 & & (0.0900) & (0.0861) & (0.0889) & \\[2pt]
\bottomrule
\end{tabular}
\end{table}
\section{Ablation Study}
\label{sec:ablation}

To isolate the effect of the training-time constraint, the same boosting model is fitted twice on every example, once with the coordinate norm bound imposed at every leaf and once without it. The coordinates, reference subspaces, cross-validation folds and settings are the same, so the constraint is the only difference between the two fits. The predicted coordinates are examined before the prediction-time safeguard is applied.

\subsection{Setup and Diagnostics}
\label{sec:exceed}
\label{sec:outside}
\label{sec:stages}

Each raw held-out prediction is checked against the conservative bound $\|\hat{\bm{y}}\|_2 < \pi/2$ and for membership of $\Omega_p$, the larger region on which the coordinate-to-subspace map is one-to-one. The raw norm is also recorded after each boosting stage. Table~\ref{tab:ablation} gives these quantities.

In every fold of every example the base score lies inside the ball, so its projection never acts, and the largest accumulated training prediction reaches $1.00$ of the bound without crossing it. Most target coordinates of the three scalar sweeps lie outside the ball, so the bound is conservative for these sweeps. On Kuramoto--Sivashinsky the largest target principal angle is $\pi/2$ to machine precision, so that target cannot be told apart from the boundary of $\Omega_p$. On the beam it is $\pi/2$ to within $10^{-6}$, and on the cylinder and Kolmogorov it is at most $0.999$ of $\pi/2$.

\begin{table}[htbp]
\centering
\caption{Ablation of the constraint on raw coordinates before the prediction-time safeguard. Past $\pi/2$ counts the coordinates whose Euclidean norm exceeds $\pi/2$. Max norm is the largest coordinate norm divided by $\pi/2$, with the enforced radius $R$ rounding to $1.00$. $\|\bm{Z}\|_2>\pi/2$ counts the coordinates whose largest singular value exceeds $\pi/2$. Stages is the mean share of boosting stages at which a held-out prediction exceeds the bound. Leaves is the share of fitted-tree nodes at which the constraint is active, excluding candidate splits. The Targets rows use one reference for the whole sweep, at the median position, while the other rows use the reference of each training fold.}
\label{tab:ablation}
\footnotesize
\setlength{\tabcolsep}{4pt}
\begin{tabular}{lccccc}
\toprule
Version & Past $\pi/2$ & Max norm$/(\pi/2)$ & $\|\bm{Z}\|_2 > \pi/2$ & Stages & Leaves \\
\midrule
\multicolumn{6}{l}{\emph{Cylinder}, $46$ parameters} \\
\quad Targets & 36/46 & 1.54 & 0 & -- & -- \\
\quad cXGBoost & 0/46 & 1.00 & 0 & 0.0\% & 85.8\% \\
\quad Unconstrained & 36/46 & 1.51 & 0 & 69.7\% & 0.0\% \\
\midrule
\multicolumn{6}{l}{\emph{Kolmogorov}, $34$ parameters} \\
\quad Targets & 33/34 & 2.21 & 0 & -- & -- \\
\quad cXGBoost & 0/34 & 1.00 & 0 & 0.0\% & 97.0\% \\
\quad Unconstrained & 33/34 & 2.09 & 0 & 91.5\% & 0.0\% \\
\midrule
\multicolumn{6}{l}{\emph{Kuramoto--Sivashinsky}, $44$ parameters} \\
\quad Targets & 43/44 & 2.46 & 0 & -- & -- \\
\quad cXGBoost & 0/44 & 1.00 & 0 & 0.0\% & 93.5\% \\
\quad Unconstrained & 41/44 & 2.34 & 8 & 86.3\% & 0.0\% \\
\midrule
\multicolumn{6}{l}{\emph{Beam}, $25$ parameters} \\
\quad Targets & 7/25 & 2.19 & 0 & -- & -- \\
\quad cXGBoost & 2/25 & 1.23 & 2 & 3.2\% & 59.4\% \\
\quad Unconstrained & 8/25 & 1.87 & 0 & 21.2\% & 0.0\% \\
\bottomrule
\end{tabular}
\end{table}

Every raw cXGBoost held-out prediction on the three scalar sweeps satisfies the bound, as Section~\ref{sec:cxgboost} guarantees for a scalar parameter. On these three the unconstrained model exceeds it at most held-out parameters, and across all four examples it leaves $\Omega_p$ only on Kuramoto--Sivashinsky. 

With the reference taken instead as the training basis with the smallest mean geodesic distance to the other training bases, the constrained model stays inside the bound at every held-out parameter of every example, and the unconstrained model exceeds it at $36$ of $46$, $28$ of $34$, $41$ of $44$ and $5$ of $25$ parameters, with two departures from $\Omega_p$ on Kuramoto--Sivashinsky and one on the beam.

At the active leaves the scaled iterate of Section~\ref{sec:cxgboost}
replaces the projection at $56\%$ of the Dykstra calls on the cylinder,
$25\%$ on Kolmogorov, $30\%$ on Kuramoto--Sivashinsky and $2.4\%$ on the
beam, candidate splits included. In one fold of each scalar sweep, every fortieth active leaf problem is re-solved by an interior-point method. The leaf objective of the scaled iterate is then at most $1.1\%$ above the constrained optimum, and at most $0.002\%$ above it at the median of each sweep. Both gaps are measured relative to how much the constraint raises the leaf objective above that of the ordinary XGBoost leaf value.

\subsection{How the Constraint Changes the Boosting Fit}
\label{sec:constraint_effect}

\begin{figure}[!htb]
    \centering
    \includegraphics[width=\linewidth]{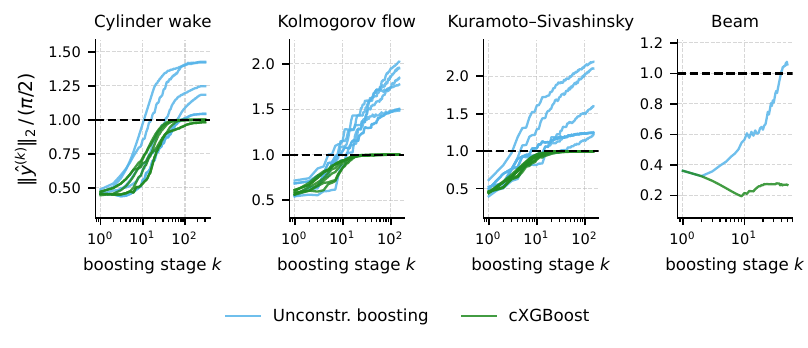}
    \caption{Norm of the raw held-out coordinate after each boosting stage for selected held-out parameters, one fold per example, as a multiple of the bound $\pi/2$.}
    \label{fig:stagenorms}
\end{figure}

The constrained and unconstrained ensembles are identical until the constraint first becomes active at a leaf or a candidate split, and can differ from that stage on. Figure~\ref{fig:stagenorms} plots the norm of the raw held-out coordinate after each boosting stage for one cross-validation fold of each example. It shows every held-out parameter of that fold whose prediction exceeds the bound at some stage in either model, with the trajectories of both models for each of them. The two beam violations of Table~\ref{tab:ablation} occur in other folds. The two sets of trajectories separate during fitting, so the constraint acts while the trees are fitted and not only on the final prediction.

\subsection{Behaviour at Unseen Parameters}
\label{sec:limits}

With two or more parameters the training-time guarantee need not extend to a new parameter, which may select a combination of leaves that no training parameter selects (Section~\ref{sec:cxgboost}). On the beam every held-out parameter
selects such a combination, and in the other three examples none does.

On the beam the constraint reduces the raw violations without removing them (Table~\ref{tab:ablation}), so the prediction-time safeguard is still required. The two constrained predictions that exceed the bound also lie outside $\Omega_p$, while no unconstrained beam prediction does. The Euclidean norm combines all $r$ singular values, whereas membership of $\Omega_p$ depends only on the largest. At these two parameters the largest singular values of the constrained predictions are $1.01$ and $1.19$ of $\pi/2$, against $0.84$ and $0.91$ for the unconstrained ones. The safeguard then projects both onto the ball before they are mapped back, so every delivered beam prediction lies inside $\Omega_p$.
\section{Conclusions}
\label{sec:conclusions}

This work introduced constrained extreme gradient boosting (cXGBoost) for
predicting parameter-dependent POD subspaces. The method represents each
subspace by the Euclidean coordinate of its Grassmann tangent vector and
incorporates the conservative coordinate norm bound directly into
vector-valued boosting. Under the squared Euclidean loss, the constrained
leaf problem reduces to a projection onto an intersection of Euclidean balls, and every accumulated prediction at a
training parameter stays inside the bound at every boosting stage.

On all four examples, cXGBoost has a lower mean cross-validated reconstruction error than Grassmann interpolation and the projected Gaussian process (pGP), and it has a lower error than the pGP at most parameters.
The ablation shows that the constraint changes the boosting trajectory once an unconstrained update would cross the bound.

The bound is enforced at the training parameters, and for a scalar parameter it also holds at every other parameter, so no held-out prediction of the three scalar sweeps needed the safeguard. With two parameters this need not hold, and on the beam $2$ of the $25$ raw held-out predictions exceed the bound and lie outside the one-to-one region before the safeguard is applied. On Kolmogorov and
Kuramoto--Sivashinsky all three methods have high absolute reconstruction
error. Future work can extend the guarantee to unseen parameters in problems with several parameters, develop coordinate representations that remain one-to-one over
wider parameter ranges and
include predictive uncertainty.

\section*{Conflict of Interest}

The authors declare that they have no known competing financial interests or
personal relationships that could have appeared to influence the work reported
in this paper.

\section*{Funding}

This research did not receive any specific grant from funding agencies in the
public, commercial or not-for-profit sectors.

\section*{Data Availability Statement}

The source code implementing cXGBoost is available under the MIT licence at
\url{https://github.com/MelikaBaghi/cXGBoost}. The result files behind all tables and plots are available in the same repository. The drivers that generate them and the simulation data, from which the field figures are drawn, are available from the authors on reasonable request.

\bibliography{ref}

\end{document}